\documentclass{sig-alternate-05-2015}
\usepackage{bm}
\usepackage{booktabs}
\usepackage{multirow}
\usepackage{tabularx}
\newcommand{\tabincell}[2]{\begin{tabular}{@{}#1@{}}#2\end{tabular}}

\makeatletter
\newcommand{\rmnum}[1]{\romannumeral #1}
\newcommand{\Rmnum}[1]{\expandafter\@slowromancap\romannumeral #1@}
\makeatother

\usepackage{makecell}
\usepackage{hhline}
\usepackage{cellspace}
\begin{document}

\setcopyright{acmcopyright}

\CopyrightYear{2016}
\setcopyright{acmcopyright}
\conferenceinfo{CIKM'16 ,}{October 24-28, 2016, Indianapolis, IN, USA}
\isbn{978-1-4503-4073-1/16/10}\acmPrice{\$15.00}
\doi{http://dx.doi.org/10.1145/2983323.2983861}





%

\title{Skipping Word: A Character-Sequential Representation based Framework for Question Answering}
%
%
%
%
%

\numberofauthors{3} 
%
\author{
\alignauthor
Lingxun Meng\\
\affaddr{Sogou Inc}\\
\affaddr{Beijing, 100084, China.}\\
\email{menglingxun@sogou-inc.com}
\alignauthor
Yan Li, Mengyi Liu\\
\affaddr{Key Lab of Intelligent Information Processing of CAS}\\
\affaddr{, ICT, Beijing, 100190, China.}\\
\email{\{yan.li, mengyi.liu\}@vipl.ict.ac.cn}
\alignauthor
Peng Shu\\
\affaddr{Sogou Inc}\\
\affaddr{Beijing, 100084, China.}\\
\email{shupeng203672@sogou-inc.com}%
}

\maketitle
\begin{abstract}
Recent works using artificial neural networks based on word distributed representation greatly boost the performance of various natural language learning tasks, especially question answering. Though, they also carry along with some attendant problems, such as corpus selection for embedding learning, dictionary transformation for different learning tasks, etc. In this paper, we propose to straightforwardly model sentences by means of character sequences, and then utilize convolutional neural networks to integrate character embedding learning together with point-wise answer selection training. Compared with deep models pre-trained on word embedding (WE) strategy, our character-sequential representation (CSR) based method shows a much simpler procedure and more stable performance across different benchmarks. Extensive experiments on two benchmark answer selection datasets exhibit the competitive performance compared with the state-of-the-art methods.
\end{abstract}


\section{Introduction}
Inspired by the achievements of convolutional networks (a.k.a, ConvNets) in the field of computer vision, more and more researchers constitute ConvNets for kinds of natural language processing tasks, e.g., text classification \cite{kim2014convolutional}, short text pair re-ranking \cite{yih2013question,severyn2015learning}, e.t.c. Nearly all of them pointed out that ConvNets could successfully capture local syntactic structure to boost performance. In addition to the appropriate use of ConvNets, researchers also owed their success to recent distributed representation of words, like {\em word2vec} \cite{mikolov2013distributed}. No doubt perfect word embedding (WE) could greatly boost the performance, but there are so many English word embedding systems available online, trained with different corpus, algorithms and preprocessing methods. Thus, evaluating the effectiveness of different word embedding systems becomes a laborious job. Moreover, for other languages, like Chinese, there exists few word embedding systems available, so this further makes the selection of right corpus, right word segmentation tools and even right algorithms more expensive and difficult. \\
\indent Recently, some researchers start to use character representation to constitute different natural language models based on the fact that character embedding model not only alleviates the parametric burden significantly, but also brings more ability of handling morphologically rich language and out-of-vocabulary words \cite{kim2015character}. Besides, character-level representation has also been successfully investigated for text classification \cite{zhang2015character}, where one-hot representation is feed into Deep ConvNets. No matter what algorithms are used to construct task-specific semantic learning framework, the same concept is shared that characters make words, and then words construct whole sentences. In fact of reality, we believe that root word is the primary lexical unit of comprehension, so direct construction from characters to sentences is reasonable and enough for semantic learning which forms our original motivation.
\\\indent In this paper, we take the first attempt to construct text matching based on character-sequential representation (CSR). Different from previous works, we first directly model sentences as sequences of characters, skipping the traditional word-level comprehension, and then utilize shallow weight-sharing ConvNets to learn the matching task. Experiments on two popular question answering benchmarks show the competitive performance compared with the other state-of-the-art methods.
\begin{figure*}
\centering
\includegraphics[width=18cm,height=7.5cm]{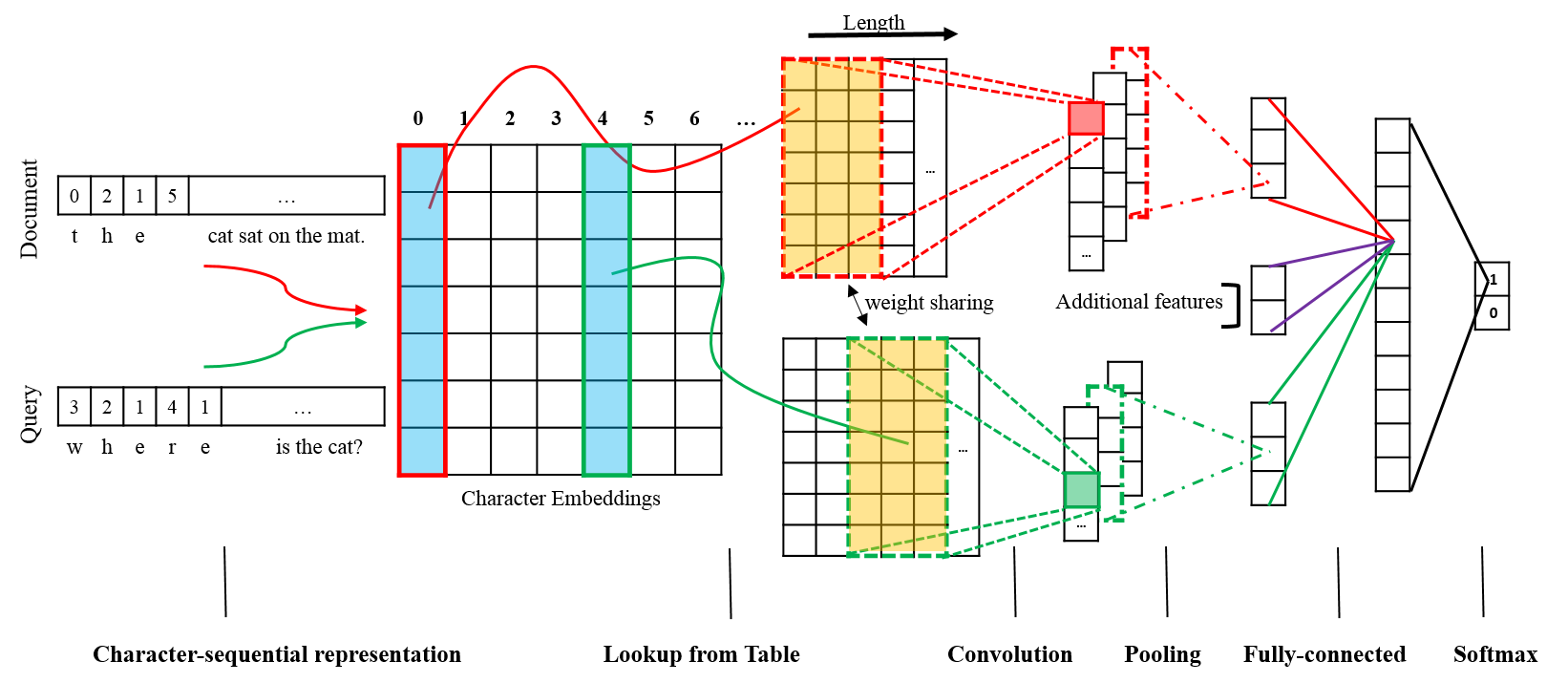}
\label{fig:thewholeframework}
\caption{Our framework for sentence-pair matching.}
\end{figure*}
\section{Semantic Match Learning} 
Text matching is generally believed as the key of many natural language processing tasks, such as question answering, information retrieval and relevance classification. The majority of previous works formulated sentence, phrase or text using semantic parsing methods, which require experts to handcraft grammars and knowledge base schema \cite{Berant:EMNLP13}. Recently, lots of works spring up by using ConvNets to constitute text based on word representation \cite{severyn2015learning}. Our method follows this constitution, but replaces the word look-up table with more fundamental character look-up table, and the look-up table, convolutional filters and normalization parameters are shared between question-answer pairs (see Fig. \ref{fig:thewholeframework} for better understanding).
\subsection{Character-Sequential Representation} 
The sentence in this paper is represented as a sequence of characters: $[c_0,...,c_{|s|-1}]$, where $c_i$ is drawn from the character set $C$. Each character is represented by its corresponding index in the character set $C$, namely \{$0, \cdots, |C|-1$\}. To learn the character representation, we design the first layer of our model as a look-up table, i.e., character embedding matrix ${\bm W} \in \mathbb{R}^{d \times| \bm C|}$, where each column represents a character embedding $\bm c$ which is $d$-dimensional. Forwarding from this layer, we get our sentence represented as follow:
\begin{equation}\bm S = [\bm c_0 \qquad \bm c_1 \qquad \cdots \qquad \bm c_{|s|-1}].\label{eq:csr}\end{equation}
\indent Different from previous character embedding methods, we construct sentences directly from characters, skipping the transitional step of modeling word-level comprehension (i.e., character-sentence rather than character-word-sentence). We think the comprehension unit is radical, so semantic learning from character representation is feasible.
\subsection{Convolution and Pooling}
The convolution layer consists of a filter bank $\bm F\in \mathbb {R}^{n \times c \times w}$, along with filter biases $\bm b \in \mathbb {R}^n$, where $n$ and $w$ refer to the number and width of filters respectively, and $c$ denotes the channels of data from the lower layer. More specifically, for the first convolution layer, $c$ equals to the embedding dimension $d$, which means to convolve across the characters to learn the pattern. In our construction, all the convolution operation is one-dimensional one. Formally, the output of the convolution with filter bank $\bm F$ over sentence $\bm S$ is computed as follow:
\begin{equation}
\bm T=[t_{i,j}]=\bm F* \bm S+\bm b=[\bm F_i^T\bm S_{j-w+1:j}+b_i]
\label {eq:1}
\end{equation}
where $i$ indexes the number of filter bank $F$, $F_i \in \mathbb {R}^{(d*w)}$ is the vectorization of each filter map,
$S_{j-w+1:j}$ vectorizes a sliding window matrix covering embeddings from the index ${(j-w+1)}^{th}$ to ${j}^{th}$.\\
\indent The above equation formulates two types of convolution: \em wide \em and \em narrow \em. The previous works \cite{DBLP:conf/acl/KalchbrennerGB14} pointed out that using \em wide \em type of convolution was able to better and more frequently reach boundaries of sentences than the \em narrow \em type. More importantly, the \em wide \em convolution operation could ignore the requisition of the \em narrow \em type that filter width must be smaller than input data width to guarantee a valid non-empty result. In our implementation, we use {\em narrow} type since our character based sequence is not sensitive to boundary information that much. However, we consist that using {\em wild} type may always guarantee a better performance. In this way, we get the output $\bm T \in \mathbb{R}^{n\times (|s|-w+1)}$.\\
\indent Pooling operation is often used to eliminate the differences in length for sentence representation, and it is also benefit to filter out unimportant or noisy characters and words. Technically, there exist two types of pooling strategy, i.e., {\em average} pooling and {\em max} pooling, and {\em max} pooling is more widely used due to its fast convergence and less computational complexity. Recently, {\em $k$-max} pooling strategy \cite{DBLP:conf/acl/KalchbrennerGB14} was proposed to extract $k$ large activation values rather than the topmost one, and this could construct a deeper architecture with several convolutional layers. For simplification, we adopt the normal {\em max} pooling strategy in this work. We get our result, pool$(\bm T)$: $\mathbb{R}^{n\times (|s|-w+1)} \rightarrow \mathbb{R}^n$.
\subsection{Batch Normalization}
Batch Normalization (BN) ~\cite{ioffe2015batch} was originally proposed to reduce the changes in distribution of each layers input during training, which was called \em internal covariance shift \em. As an important component of deep networks, it allows us to use much higher learning rates and be less careful about initial data preprocessing and weights initialization. Furthermore, it can also act as a regularization, in some cases eliminating the use of dropout.\\
\indent Specifically, for the given output $\bm T$ after the convolution operation, we will normalize each ${\bm t}_i$ along the filter number axis as follow:
\begin{equation} {\hat{\bm{t}_i}}=\frac{\bm{t}_i-\textrm{E}[\bm{t}_i ]}{\sqrt{\textrm{Var}[\bm{t}_i]}},\label{eq:2}\end{equation} where the expectation and variance are computed over the whole training data set. A pair of parameters $\bm{\gamma}_i$ and $\bm{\beta}_i$ (with the same dimension with $\bm{t}_i$) are further used to scale and shift the normalized value as follow: \begin{equation} {\textrm{BN}}(\bm{T})=[\bm{\gamma}_i\cdot {\hat{\bm{t}_i}}+ \bm{\beta}_i].\label{eq:3}\end{equation}
Above parameters are learnt along with the whole model training, and it usually suggests that Batch Normalization should better be used before nonlinearity activation and covers both fully connected layers and convolutional layers.
\subsection{{\secit Pointwise} Learning to Rank}
\label{ssec:pwrl}
Since our character-sequential model focuses on binary classification task, e.g., question answering, we deploy the cross-entropy loss to discriminatively train our framework as follow:\begin{equation}L_{\theta}=-\frac {1}{N}\sum_{i=1}^N\sum_{j=1}^M(y_j^{(i)}\log p_j^{(i)})+\lambda \Arrowvert \bm F \Arrowvert^2\label{eq:4}\end{equation}
where $p_j^{(i)}$ is the $j^{th}$ probability output of sample $i^{th}$ through our networks, $y_j^{(i)}$ is the corresponding ground truth, $\theta$ contains all the parameters, i.e., $\theta =\{{\bm W}, {\bm F}, {\bm b}, {\bm \gamma}, {\bm \beta}\}$, $\lambda$ is set to be $5e^{-4}$.\\
\indent We use Stochastic Gradient Descent (SGD) to optimize our network, and  AdaDelta is used to automatically adapt the learning rate during the training procedure. For higher performance, hyper-parameter selection is conducted on the development set, and BN layer after each convolution layer is also added to speed up the network optimization. In addition, dropout is applied after the first hidden layer for regularization, and early stopping is used to prevent over fitting with a patience of 5 epochs.
\begin{table}
\centering
\small
\begin{tabular}{ccccc}
\toprule[1.5pt]
{\bf Dataset} & {\bf Set} & {\bf \#Question} & {\bf \#QApairs} & {\bf \%Corect}\\
\midrule[0.8pt]
\multirow{3}{*}{{TrecQA}}
& Train & {94} & {4,718} & {7.4\%} \\
& Dev & {65} & {1,117} & {18.4\%} \\
& Test & {68} & {1,442} & {17.2\%} \\
\midrule[0.8pt]
\multirow{3}{*}{{WikiQA}}
& Train & {2,118} & {20,360} & {5.11\%} \\
& Dev & {296} & {2,733} & {5.12\%} \\
& Test & {633} & {6,165} & {4.75\%} \\
\bottomrule[1.5pt]
\end{tabular}
\caption{Statistics of the answer selection datasets.}
\label{tab:datasetsinfo}
\end{table}
\section{Experiments}
We evaluate the proposed method on two benchmarks of answer selection problem, i.e., TrecQA \cite{wang2007jeopardy} and WikiQA \cite{yang2015wikiqa}.
\subsection{Datasets}
\label{sec:dataset}
TrecQA\footnote{\url{http://cs.stanford.edu/people/mengqiu/data/qg-emnlp07-data.tgz}} was collected from Text Retrieval Conference (TREC) QA track (8-13) data, and the correctness of the selected answer was guaranteed by manual judgment for parts of the dataset. \\
\indent WikiQA\footnote{\url{http://research.microsoft.com/en-us/downloads/4495da01-db8c-4041-a7f6-7984a4f6a905/}} is another open domain question-answer dataset, collected from real queries of Bing search engine without human editorial revision.  Table \ref {tab:datasetsinfo} summarizes the statistics on the two datasets.
\begin{table}
\centering
\begin{tabular}{|l|c|c|}
\hline
{\bf NetConfig} & {\bf WE} & {\bf CSR}\\\hline
Embedding dim & 50, 300 & 50\\\hline
CNN filter width & 5 & 3, 5\\\hline
CNN num. of filters & 100 & 128, 32\\\hline
Dropout rate & 0.5& 0\\\hline
Batch Norm & w & w, w/o\\\hline
Sim Matrix & w & w/o\\\hline
Pooling & $average$ & $max$\\\hline
\end{tabular}
\caption{The configuration of networks used for WE and CSR based networks in the TrecQA and WikiQA dataset, respectively.}
\label{tab:netconfig}
\end{table}
\subsection{Experimental Setup}
\label{sec:setup}
We set the maximum word length of questions and answers to be 30/80 for TrecQA dataset, and 50/50 for WikiQA dataset respectively. Alike, we set the maximum character length of questions and answers to be 192/386 for TrecQA dataset, and 125/386 for WikiQA dataset respectively. All the words can not be found in the pre-trained word look-up table indexed by an extra index named {\em \small UNK\_IND}.\\
\indent Following \cite{zhang2015character}, our character set $C$ also consists of 71 characters, including 26 english lower letters, 10 digits, 33 other characters, the new line character, the padding symbol and the unknown symbol. The characters are as follows:
\begin{center}
{\url{abcdefghijklmnopqrstuvwxyz0123456789}
\url{,;.!?:'"/\|_@#$%~^`&*+-=<>()[]{}}}
\end{center}
\section{Results and Discussions}
\label{sec:rst}
The configuration parameters of both word embedding (WE) and character-sequential representation (CSR) based networks are listed in Table \ref{tab:netconfig} (we carefully implemented WE based network architecture according to \cite{severyn2015learning}). Additionally, similarity weight matrix ($sim(x, y) = x^T {\bf M} y$, where ${\bf M}$ is the similarity matrix) between question answer pairs is learnt along with WE models training, that is mentioned in \cite{severyn2015learning} to perform a better matching. For fair comparison, the simple word-level overlap feature and IDF-weighted word overlap feature are used in both WE and CSR based models.\\
\indent We use the open deep learning framework Caffe \cite{jia2014caffe} to complete our implementation. Due to the randomness caused by $max$-pooling implementation in Caffe, we run 10 times the training and report the mean and variance value as our final result.\\
\indent For quantitative evaluation, we use two classic measurements, i.e., Mean Average Precision (MAP) and Mean Reciprocal Rank (MRR), which are consistent with previous works. Specifically, the official {\em trec\_eval} scorer tool is used to compute the above measurements\footnote{\url{http://trec.nist.gov/trec_eval/}}.
\begin{table}
\centering
\begin{tabular}{ccccc}
\toprule[1.51pt]
\multirow{2}{*}{{Model}} &
\multicolumn{2}{c}{TrecQA}  & \multicolumn{2}{c}{WikiQA}\\
\cmidrule[1pt]{2-5}
  & MAP & MRR & MAP & MRR\\
\midrule[1pt]
WE(glv-t) & .6719 & .7702 & .6445 & .6583\\
WE(glv-6) & .6999 & .8048 & .6430 & .6542\\
WE(w2v) & .5898 & .6804  & .6533 & .6693\\
\midrule[0.6pt]
CSR & \tabincell{c} {\bf .7295 \\ {\small $\pm$ .0036}} & \tabincell{c}{\bf .8232 \\ {\small $\pm$ .0031}} & \tabincell{c}{\bf .6608 \\ {\small $\pm$ .0099}} & \tabincell{c}{\bf .6811 \\ {\small $\pm$ .0101}}\\
\bottomrule[1.5pt]
\end{tabular}
\caption{Results of WE models based on different pre-trained WE and CSR models. w2v: word2vec. glv-t: GloVe-twitter. glv-6: GloVe-6B.}
\label{tab:rst1}
\end{table}
\\\indent We use three sets of word embedding for WE based models: (\rmnum{1}) {\bf word2vec}\footnote{\url{https://code.google.com/archive/p/word2vec/}} \cite{mikolov2013distributed} is trained on part of Google News dataset, containing 100 billion words; (\rmnum{2}) {\bf GloVe-twitter} \cite{pennington2014glove} is trained on 2 billion tweets, containing 27 billion tokens; (\rmnum{3}) {\bf GloVe-6B}\footnote{\url{http://nlp.stanford.edu/projects/glove/}} is trained on Wikipedia data and the fifth English Gigawords with totally 6 Billion tokens. The dimension of our character embedding for both datasets is set to 50, and standard setup is adopted of considering questions that have both correct and negative answers for evaluation. Table \ref{tab:rst1} shows the results of three versions of WE models, and please kindly note that our WE based model implementations are all slightly better than the original paper reports. We can also see that different pre-trained WE model results in unstable performance even in the same dataset, especially TrecQA, and the same embedding tool shows totally different performance on different datasets. Although our CSR based model adopts end-to-end strategy to learn the representation directly from the data without extra corpus, it exhibits competitive result in both datasets.
\begin{table}
\begin{small}
\centering
\begin{tabular}{@{}lc@{}clc@{}c@{}}
\toprule[1.5pt]
\multirow{2}{*}{{ Model}} &
\multicolumn{2}{c}{TrecQA} & \multirow{2}{*}{{Model}} & \multicolumn{2}{c}{WikiQA}\\
\cmidrule[1pt]{2-3} \cmidrule[1pt]{5-6}
  & MAP & MRR & & MAP & MRR\\
\midrule[1pt]
Jeopardy\cite{wang2007jeopardy} & .6029 & .6852 & PV\cite{le2014distributed} &  .5976 & .6058\\
LCLR\cite{yih2013question} & .7092 & .7700 & Bigram\cite{yu2014deep} & .5993 & .6086\\
Bigram\cite{yu2014deep} & .7113 & .7846 & LCLR\cite{yih2013question} & .6190 & .6190\\
NASM \cite{miao2015neural}& {\bf .7339} & .8117  & CNN\cite{yang2015wikiqa}& .6520 & .6652\\
\midrule[0.6pt]
CSR & \tabincell{c}{ .7295 \\ {\small $\pm$ .0036}} &\tabincell{c}{\bf .8232 \\ {\small $\pm$ .0031}} &CSR&  \tabincell{c}{\bf .6608 \\ {\small $\pm$ .0099}} & \tabincell{c}{\bf .6811 \\ {\small $\pm$ .0101}}\\
\bottomrule[1.5pt]
\end{tabular}
\caption{Results of our models and other methods from the literature.}
\label{tab:rst2}
\end{small}
\end{table}
\\\indent Table \ref{tab:rst2} compares the proposed model with the previous state-of-the-art methods. Different from the comparative methods using noisy sampled TRAIN-ALL as training set, we report our final results trained on the TRAIN set, which is much smaller than TRAIN-ALL. It is obvious that the proposed CSR exhibits its effectiveness, even without any extra-corpus for embedding pre-training.
\section {Conclusion}
\label{sect:ccls}
In this paper, we address the problem of word embedding (WE) based answer selection that extra-corpus are needed for perfect word embedding. To solve this problem, we propose an end-to-end character-sequential representation (CSR) based ConvNets for answer selection, which simply models sentences by character sequences, and then integrates character embedding learning with point-wise answer ranking training. Extensive experiments provide further evidence that construction directly from characters to sentences works as well as or even better than the character-word-sentence hierarchical formula. We hope this work can bring some insights to the research and industrial communities that corpus-free and skip-word representation from character directly to sentence as an end-to-end strategy could play its role on various languages with its advantages of simplification and concise body.

\section{Acknowledgments}
We thank all the co-workers in wireless sponsored search advertising (ADWR) team for details discussion and consummation. The authors also thank Dingding Qian and Kai Chen for the paper submission suggestion. This paper is partially supported by Natural Science Foundation of Zhejiang Province, P.R.China under contract No.LQ15F020005.
%
\bibliographystyle{abbrv}
\bibliography{sigproc}  
%
%

\end{document}